\documentclass[applsci,article,accept,moreauthors,pdftex]{mdpi} 

\usepackage{mathtools}
\usepackage{amsmath}
\firstpage{1} 
\pubvolume{xx}
\issuenum{1}
\articlenumber{5}
\pubyear{2019}
\copyrightyear{2019}
\history{Received: date; Accepted: date; Published: date}

\Title{Multi-Task Learning for Multi-Dimensional Regression: Application to Luminescence Sensing}

\Author{Umberto, Michelucci $^{1}$, Francesca Venturini $^{1,2}$}

\AuthorNames{Umberto Michelucci, Francesca Venturini}

\address{%
$^{1}$ \quad TOELT LLC; umberto.michelucci@toelt.ai\\
$^{2}$ \quad Zürich University of Applied Science, ; francesca.venturini@zhaw.ch}

\corres{Correspondence: umberto.michelucci@toelt.ai}

\abstract{
The classical approach to non-linear regression in physics, is to take a mathematical model describing the functional dependence of  the dependent variable from a set of independent variables, and then, using non-linear fitting algorithms, extract the parameters used in the modeling. Particularly challenging are real systems, characterized by several additional influencing factors related to specific components, like electronics or optical parts. In such cases, to make the model reproduce the data, empirically determined terms are built-in the models to compensate for the impossibility of modeling things that are, by construction, impossible to model. A new approach to solve this issue is to use neural networks, particularly feed-forward architectures with a sufficient number of hidden layers and an appropriate number of output neurons, each responsible for predicting the desired variables.
Unfortunately, feed-forward neural networks (FFNNs) usually perform less efficiently when applied to multi-dimensional regression problems, that is when they are required to predict simultaneously multiple variables that depend from the input dataset in fundamentally different ways. 
To address this problem, we propose multi-task learning (MTL) architectures. These are characterized by multiple branches of task-specific layers, which have as input the output of a common set of layers.
To demonstrate the power of this approach for multi-dimensional regression, the method is applied to luminescence sensing. Here the MTL architecture allows predicting multiple parameters, the oxygen concentration and the temperature, from a single set of measurements.}


\featuredapplication{Multi-task learning; multi-parameter luminesce sensing.}

\begin{document}

\section{Introduction}

The classical use of regression in physics, sometimes also referred to as non-linear fitting, is to try to determine $d$ quantities ${\mathbold y} \in \mathbb{R}^d $ from a set of $n$ measurements ${\mathbold x} \in \mathbb{R}^{q}$ with $q \in \mathbb{N}$, using a theoretical mathematical model ${\mathbold y}= f({\mathbold x}, {\mathbold w})$ that depends on a certain number $p$ of parameters ${\mathbold w} \in \mathbb{R}^p$. Typically this is achieved by choosing the parameters ${\mathbold w}$ to minimize a selected error function, like the mean square error (MSE), with specific algorithms. To find the best solution for $f$ is a classical optimization problem \cite{nocedal, Gareth1988, Boyd2004}. 
This method, however, fails to deliver stable and accurate results, for example, when the quantities $y_i$ with $i =1,...,d$ have different physical meanings and, consequently, depend on different components of the parameter vector ${\mathbold w}$ in fundamentally distinct ways. As a result, the mathematical model may be an insufficient approximation, may be too complex for a stable implementation or may be simply unknown \cite{Gareth1988}. 

An example where the usual multi-dimensional regression approach fails is in the determination of a substance from changes in its luminescence when several environmental conditions vary in an unknown and uncontrolled way. Luminescence quenching for oxygen detection represents a widespread application relevant in many fields like biomedical imaging, environmental monitoring, or process control \cite{Borisov2018} (see Section \ref{Luminescence} for details). In this application, the quantity of interest is the concentration of molecular oxygen $[O_2]$. The measured quantity, either the luminescence intensity or luminescence intensity decay time of a special molecule (luminophore), is however equally strongly dependent on the concentration $[O_2]$ and the temperature $T$. As a result, it is not possible to extract two different physical quantities, namely $[O_2]$ and $T$, from the same set of data. Usually, $T$ is measured separately with another device and given as an input to a mathematical model describing the dependency of those two quantities from the input data. The complexity increases further if more than one luminophore is present, and several parameters (e.g. $[O_2]$, $[CO_2]$, $pH$) have to be determined \cite{Baleizao2008,Collier2011,Perez2011,Lam2011,Borisov2011}.
 
A possible method, which recently attracted great interest, is the use of feed-forward neural network (FFNN) architectures, with a certain number of hidden layers and an appropriate number of output neurons, each responsible for predicting the desired variables $y_i$ with $i=1,...,d$. In the example of oxygen sensing, the output layer would have a neuron for the oxygen concentration $[O_2]$ and one for the temperature $T$.
This work shows that, since the output neurons must use the same features (the output of the last hidden layer) for all variables \cite{Thung2018, Zhang}, FFNNs are insufficiently flexible. For the cases when the variables depend in fundamentally different ways from the inputs this approach will give a result that is at best acceptable, and at worst unusable.

This work proposes a new approach, which is based on multi-task learning (MTL) neural networks architectures. This type of architectures are characterized by multiple branches of layers, that get their input from a common set of layers. This type of networks can improve the model prediction performance by jointly learning correlated tasks \cite{Caruana1997, Thung2018, Zhang, Thrun1996, Baxter2000}. In particular, the proposed MTL architectures are applied to the problem of luminescence quenching for oxygen sensing. Their performance in the prediction of oxygen concentration and temperature is analyzed and compared to that of a classical feed-forward neural network.

In general, the proposed MTL approach may be of particular relevance in all those cases, where the mathematical model ${\mathbold y}= f({\mathbold x}, {\mathbold w})$ is not known, too complex or not really of interest and the only goal of the regression problem is to build a system that is able to determine ${\mathbold y}$ as accurately as possible.

The paper is organized as follows: Section \ref{MTL} describes non-linear regression and MTL with neural networks. Section \ref{Implementation} describes the implementation of MTL and the different neural network studied in this work. Section \ref{Luminescence} reviews luminescence quenching for oxygen sensing. The results are discussed in Section \ref{Results}.

\section{Theoretical Background}
\label{MTL}

This section briefly reviews the theoretical justification for non-linear regression with neural networks, as well as the multi-task learning approach implemented in this work.

\subsection{Neural Networks for Non-Linear Regression Problems}

In general, a neural network model is always composed of three parts \cite{Michelucci2017}:
\begin{itemize}
\item network architecture (number of layers, activation functions, etc.),
\item cost function,
\item optimizer (a method or algorithm used to minimize the cost function).
\end{itemize}
The neural networks considered in this work have a feed-forward architecture, as it is typical in regression problems. The details of the networks are described in Section \ref{Implementation}.
The cost function needs to be chosen depending on the problem to be solved. For example, the cross-entropy is a common choice when solving classification problems \cite{Michelucci2017}. 
For regression problems, as the one studied in this work, the most common cost function is the mean square error (MSE), which is defined as
\begin{equation}
MSE = \frac{1}{n}\sum_{j=1}^n \sum_{k=1}^d (y_k^{[j]}-\hat y_k^{[j]})^2
\label{MSE}
\end{equation}
where $n$ is the number of observations in the input dataset; ${\mathbold y}^{[j]} \in \mathbb{R}^d$ is the measured value of the desired quantity for the $j^{th}$ observation (indicated as a superscript between square brackets), with $j=1, ..., n$; $ \hat {\mathbold y}^{[j]} \in \mathbb{R}^d$ is the output of the network, when evaluated on the $j^{th}$ observation.
The optimizer affects the learning performance of the network but does not determine the type of problems the network can solve and therefore will not be discussed here.

A regression problem consists in minimizing the cost function, in this case the MSE (Equation \ref{MSE}), with respect to the learnable parameters of the network, which are defined in the architecture. The implicit assumption done is that there is an underlying albeit unknown function that describes the relationship between the ${\mathbold y}^{[j]}$ and the input observations (the measurements ${\mathbold x}^{[j]}$). Assuming its existence, the neural networks try to approximate it, by composing a big number of non-linear functions.
This approach relies on the implicit assumption that a network can approximate any function. For FFNN this assumption is legitimate since it was  proved mathematically \cite{Irie1988, Hornik1990, Cybenko89, Hanin2017, Lu2017, Rojas, bishop, Sprecher64}. 
This mathematical proof thus justifies the use of neural networks for regression problems. Unfortunately, not being a constructive proof, it provides neither the number of layers nor the number of neurons per layer needed to approximate this unknown function. It just tells that, with enough neurons, a neural network is able to approximate any function.

\subsection{Multi-Task Learning}

Multi-task learning is a machine learning techniques in which $n_T$ learning tasks are solved at the same time, using commonalities and differences across tasks. This approach may results in improved learning efficiency and prediction accuracy \cite{Baxter2000,Thrun1996,Caruana1997,Argyriou2006}, although the possibility of improvement depends on how information is encoded in the data. In this work MTL will be applied, for the first time, to luminescence sensing, where the luminescence data are dependent on two quantities, oxygen concentrations and temperature, which are otherwise hard to extract separately.

\begin{figure}[hbt]
\centering
\includegraphics[width=10 cm]{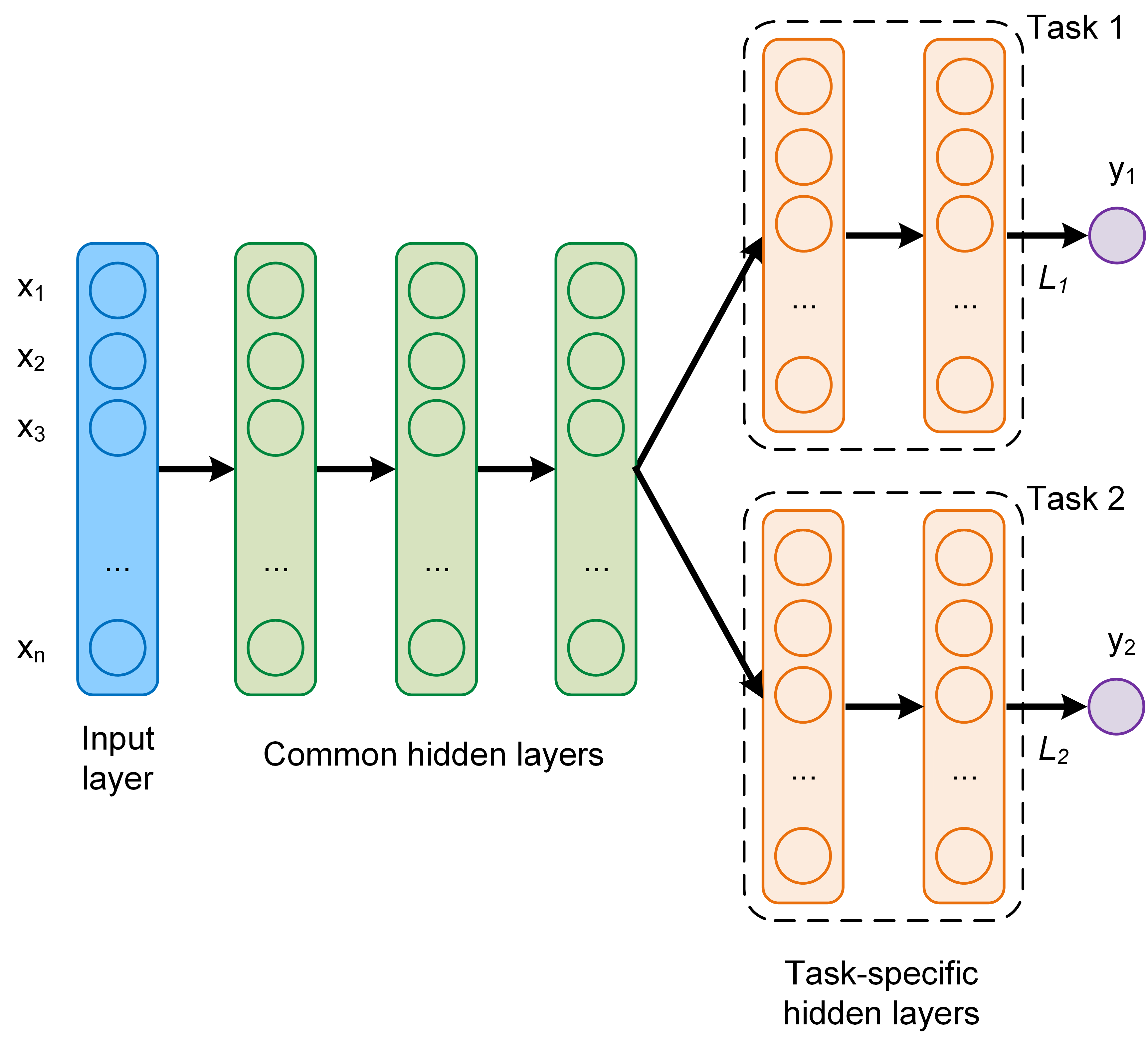}
\caption{Example of a MTL network architecture with two tasks and two outputs.}
\label{NN_MTL}
\end{figure}

An example of a simple MTL network architecture, which reflects the architectures later used in the paper, is shown in Figure \ref{NN_MTL}. This network consists of a series of common hidden layers, followed by two branches ($n_T=2$) each consisting of several task-specific hidden layers. 

The layers marked in Figure \ref{NN_MTL} as "common hidden layers" generate an output, that is typically called a "shared representation". The name comes from the fact that the output of those layers is used to evaluate both $y_1$ and $y_2$. The shared representation is then the input of a set of "task-specific hidden layers", that learn how to predict $y_1$ and $y_2$ better. Note how the common hidden layers are shared with both the tasks of predicting $y_1$ and $y_2$, while the task-specific hidden layers are specific to each task separately. The MTL network of Figure \ref{NN_MTL} uses the common hidden layers to find common features beneficial to each of the two tasks. During the training phase, learning to predict $y_1$ will influence the common hidden layers and therefore, the prediction of $y_2$, and vice-versa. A set of task-specific hidden layers will then learn specific features to each output and therefore improve the prediction accuracy.
The implicit assumption here is that the tasks have something in common; otherwise this approach will not produce the desired result.

Multiple cost functions $L_i$ with $i=1,...,n_T$, with $n_T$ the number of tasks, are required to use this network architecture. In the training phase a global cost function $L$, defined as a linear combination of the task-specific cost functions with weights $\alpha_i$ will be minimized
\begin{equation}
L = \sum_{i=1}^{n_T}\alpha_i L_i .
\label{globalcf}
\end{equation}
The parameters $\alpha_i$ have to be determined during the hyper-parameter tuning phase to optimize the network predictions.
In this paper, being the cost function the MSE (Equation \ref{MSE}), the global cost function of Equation \ref{globalcf} is
\begin{equation}
L = \sum_{i=1}^{n_T}\alpha_i \frac{1}{n}\sum_{j=1}^n \sum_{k=1}^d (y_k^{[j]}-\hat y_k^{[j]})^2
\end{equation}
where  $n_T$ is the number of tasks; $n$ is the number of observations in the input dataset; ${\mathbold y}^{[j]} \in \mathbb{R}^d$ is the measured value of the desired quantity for observation $j$, with $j=1, ..., n$; $ \hat {\mathbold y}^{[j]} \in \mathbb{R}^d$ is the output of the network, when evaluated on the $j^{th}$ observation.

\section{Neural Network Architectures and Implementation}
\label{Implementation}

In this paper three architectures, one classical FFNN and two MTL, were investigated and compared in the simultaneous prediction of oxygen concentration and temperature. To make the comparison meaningful, the parameters, which are not architecture-specific, were not varied. The details of the architectures are described in the next subsections.

In the three architectures investigated the sigmoid activation functions was used for all the neurons
\begin{equation}
\sigma(z) = \frac{1}{1+e^{-z}}.
\end{equation}
All the results were obtained with a training of 4000 epochs. The target variables ${\mathbold y}$ were normalized to vary between 0 and 1. Thus, the sigmoid activation function was used also for the output neurons $y_1$ and $y_2$. 
The input measurement, as will be explained in detail in Section \ref{Luminescence}, is a vector in $\mathbb{R}^{q}$ with $q=16$.

To minimize the cost function, the optimizer Adaptive Moment Estimation (Adam) \cite{Kingma2014, Michelucci2017} was used. The training was performed with a starting learning rate of $10^{-3}$ and using batch-learning, which means that the weights were updated only after the entire training dataset has been fed to the network. Batch-learning was chosen because of its stability and speed since it reduces the training time of a few orders of magnitude in comparison to, for example, stochastic gradient descent \cite{Michelucci2017}. Therefore it makes experimenting with different networks a feasible endeavor. 
The implementation was performed using the TensorFlow$\texttrademark$ library.

\subsection{Network A}
\label{NetworkA}

\begin{figure}[hbt]
\centering
\includegraphics[width=7 cm]{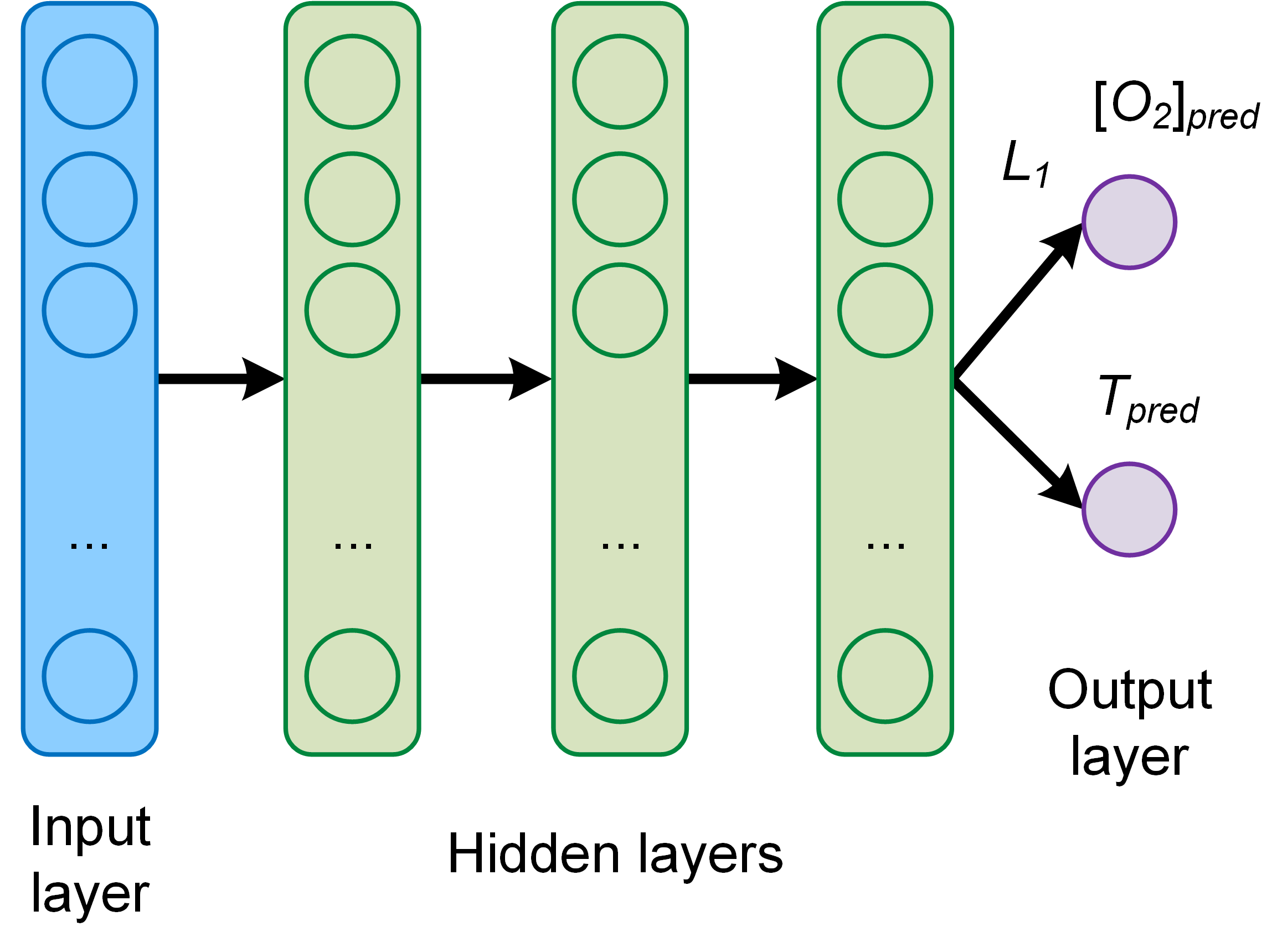}
\caption{Architecture of the feed-forward network A.}
\label{NN_simple}
\end{figure}

The first type of neural network investigated has a classical feed-forward architecture, consisting of an input layer, three hidden layers, and an output layer with two neurons $[O_2]_{pred}$ and $T_{pred}$. This architecture, labeled here as Network A, is schematically shown in Figure \ref{NN_simple}. The number of neurons of each hidden layer $n_i=\hat n$ is the same.
Each neuron in each layer gets as input the output of all neurons in the previous layer, and feeds its output to each neuron in the subsequent layer. To test the performance network A hyperparameter tuning was performed by varying the number of neurons in each of the four layers ($\hat n$). The number of neurons that was tested is $\hat n=10, 30, 50, 80$. Additional hyperparameters, like the learning rate, were not optimized and the mentioned values were kept constant.

\subsection{Network B}
\label{NetworkB}

\begin{figure}[hbt]
\centering
\includegraphics[width=9 cm]{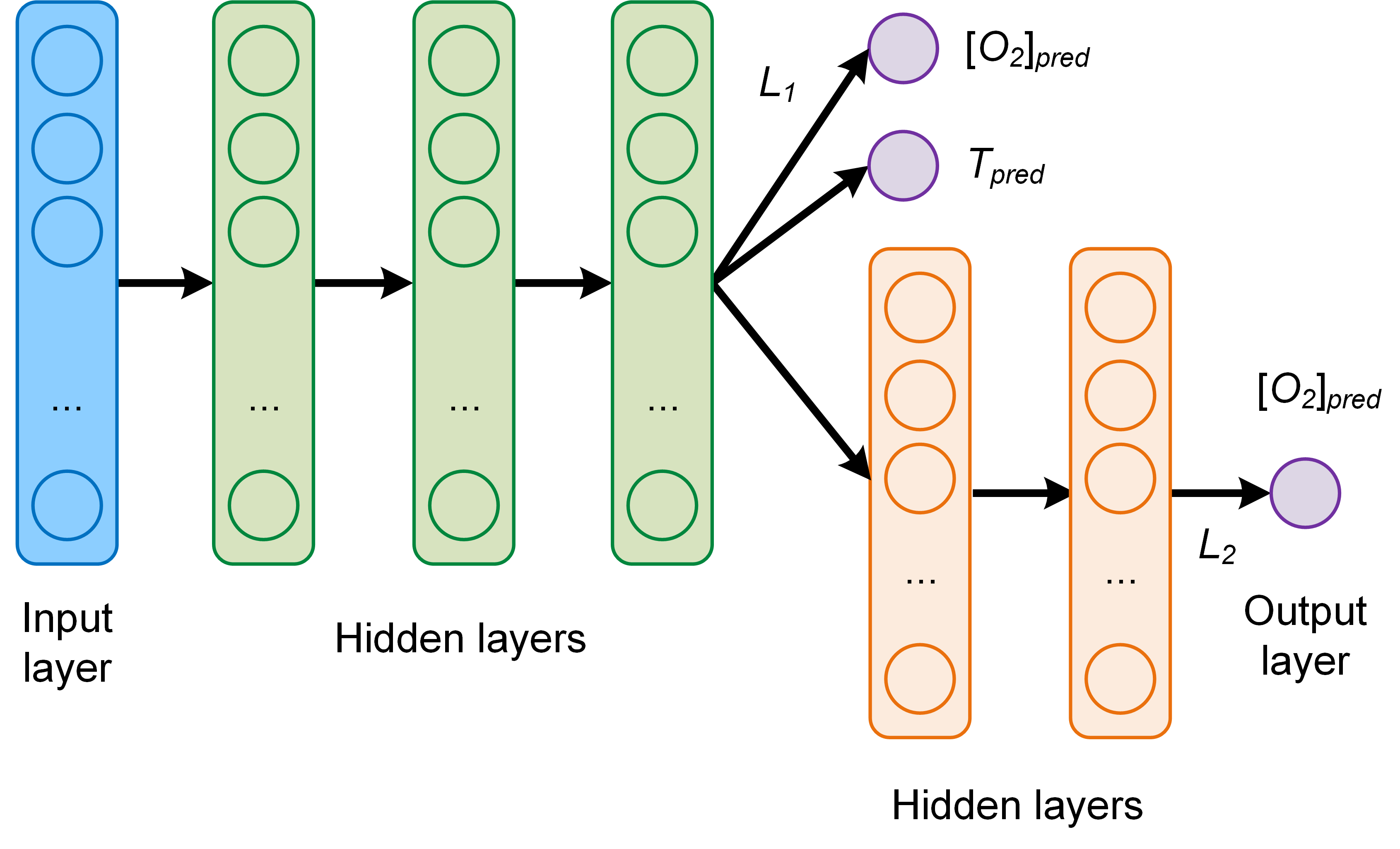}
\caption{Architecture of the feed-forward MTL network B.}
\label{NN_MTL_B}
\end{figure}

The first MTL network studied is depicted in Figure \ref{NN_MTL_B}. It consists of three common hidden layers with 50 neurons each, followed by two branches, one with two additional task-specific hidden layers used to predict $[O_2]$, and one branch without hidden layers used to predict both $[O_2]$ and $T$ at the same time. The number of neurons of each task-specific hidden layer is 5. The idea behind this network is to have a system that learns to predict  $[O_2]$ well, thanks to the further task-specific layers. The predicted $T$ is not expected to be exceptionally good since the common hidden layers must learn to predict $[O_2]_{pred}$ and $T_{pred}$ at the same time. This architecture can be of applied when one of the outputs $y_i$, here $[O_2]$, needs to be predicted with higher accuracy than the other ones.
For this network, the global cost function weights used were $\alpha_1 = 0.3$ and $\alpha_2 = 5$.

\subsection{Network C}
\label{NetworkC}

\begin{figure}[hbt]
\centering
\includegraphics[width=9 cm]{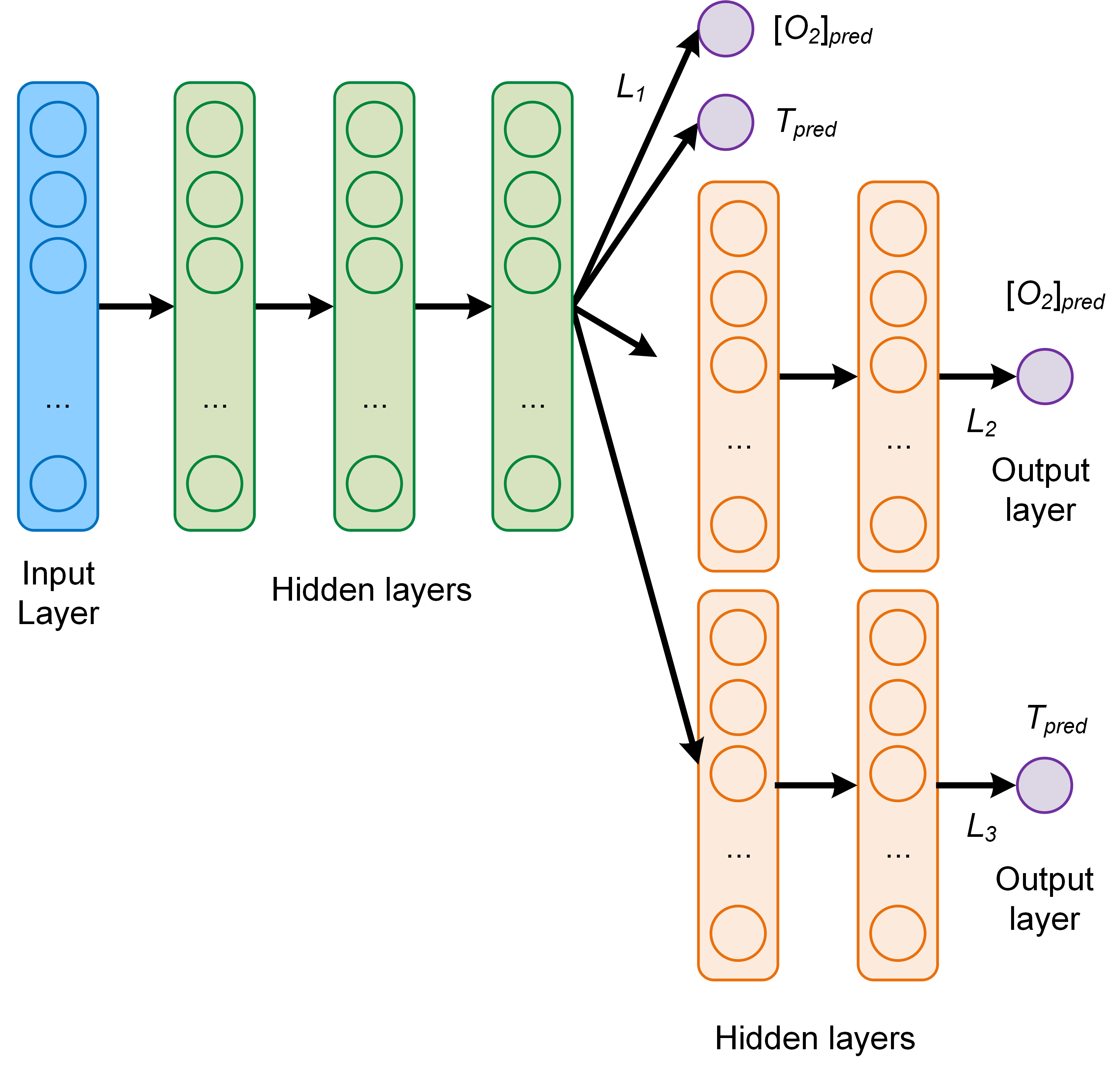}
\caption{Architecture of the feed-forward MTL network C.}
\label{NN_MTL_C}
\end{figure}

The last MTL network, depicted in Figure \ref{NN_MTL_C}, consists again of three common hidden layers with 50 neurons each, followed by three branches, two with each two additional task-specific layers to predict respectively $[O_2]$ and $T$, and then one without additional layers to predict $[O_2]$ and $T$ at the same time. The number of neurons of each task-specific hidden layer is 5, as in the network B. The global cost function weights used for the plots were $\alpha_1 = 0.3$, $\alpha_2 = 5$ and $\alpha_3 = 1$. Those values were chosen because they result in the lowest $MAE$s (see discussion in Section \ref{Results}). 

This network is of interest because of the additional task-specific layers, which are expected to improve the ability of predicting the temperature compared to the network B.

\subsection{Metrics}

The metric used to compare results from different network models is the 
absolute error ($AE$) defined as the absolute value of the difference between the predicted and the expected value for a given observation. For the oxygen concentration of the 
$j^{th}$ observation $[O_2]^{[j]}$  the $AE$ is 
\begin{equation}
\label{AE}
AE^{[j]}_{[O_2]} = |[O_2]^{[j]}_{pred}-[O_2]^{[j]}_{meas}|.
\end{equation}

The further quantity used to analyze the performance of the network is the mean absolute error ($MAE$), defined as the average of the absolute value of the difference between the predicted and the expected oxygen concentration or temperature. For example, for the oxygen prediction using the training dataset $S_{train}$, $MAE_{[O_2]}$ is defined as 
\begin{equation}
\label{MAE}
MAE_{[O_2]}(S_{train}) = \frac{1}{|S_{train}|} \sum_{j \in S_{train}}|[O_2]_{pred}^{[j]}-[O_2]_{real}^{[j]}|
\end{equation}
where $|S_{train}|$ is the size (or cardinality) of the training dataset. For example, in this work $|S_{train}|$=20000.
The $AE_{T}$ and $MAE_T$ are similarly defined.

\section{Luminescence Quenching for Oxygen and Temperature Sensing}
\label{Luminescence}

To demonstrate its advantages, the MTL approach was applied to the simultaneous determination of the oxygen concentration and temperature of a medium. There are different optical methods used to determine oxygen concentration since this is of great relevance for numerous research and application fields, ranging from biomedical imaging, packaging, environmental monitoring, process control, and chemical industry, to mention only a few \cite{Wang2014}. Among the optical methods, a well-known approach is based on luminescence quenching \cite{Demas1999,Lakowicz2006,Wolfbeis2004}.

The measuring principle is based on the quenching of the luminescence of a specific molecule (luminophore) by oxygen molecules. Because of the collisions of the luminophore with oxygen, both the luminescence intensity and decay time are reduced. 
Sensors based on this principle rely on approximate empirical models to parametrize the dependence of the sensing quantity (e.g., luminescence intensity or intensity decay time) on influencing factors. The most relevant parameter, which can be a major source of error in sensors based on luminescence sensing, is the temperature of the luminophore, since both the luminescence and the quenching phenomena are strongly dependent on temperature  \cite{Wang2014}.

The conventional approach consists in relating the change of the luminescence decay time from the oxygen concentration through a multi-parametric model, called Stern–Volmer equation \cite{Lakowicz2006}. The value of the device-specific constants is then determined through calibration. The decay time can be easily measured by modulating the intensity of the excitation. The emitted luminescence is also modulated but shows a phase shift $\theta$ which depends on the decay time. Without going into the details of the analytical model, the measured quantity, the phase shift $\theta$, is most frequently related to the oxygen concentration $\left[O_2\right]$ and temperature $T$ through the approximate equation \cite{Michelucci2019}
\begin{equation}
\frac{\tan \theta (\omega, T, [O_2])}{\tan \theta_0 (\omega, T)} = \bigg( \frac{f (\omega , T) }{1+K_{SV1} (\omega , T) \cdot \left[O_2\right]}+\frac{1-f (\omega , T) }{1+K_{SV2} (\omega , T) \cdot \left[O_2\right]} \bigg)^{-1}
\label{theta_full}
\end{equation}
where $\theta_0$ and $\theta$, respectively, are the phase shifts in the absence and presence of oxygen, $f$ and $1-f$ indicate the fraction of the total emission of two components under unquenched conditions, $K_{SV1}$ and $K_{SV2}$ are associated (Stern–Volmer) constants for each component. Since the phenomena of luminescence and luminescence quenching are strongly influenced by the temperature, the parameters $\theta_0$, $K_{SV1}$, $K_{SV2}$, and $f$ need to be modelled through different temperature dependencies \cite{Michelucci2019}. The value of the parametrisation quantities is determined through non-linear regression. $\omega$ is the angular frequency of the modulation of the excitation light. Finally,  Equation (\ref{theta_full}) must be inverted to obtain $\left[O_2\right]$ as a function of $\theta$, T, and $\omega$. To be able to have more information as input to our network, we will not use a single $\omega$ frequency value, but 16. Let's define
\begin{equation}
\label{meas}
r (\omega, T, [O_2])\equiv \frac{\tan \theta (\omega, T, [O_2])}{\tan \theta (\omega, T, [O_2]=0) }.
\end{equation}
The goal of the network is to predict the oxygen concentration and temperature from an array of values of $r (\omega, T, [O_2])$ evaluated at a discrete set of sixteen $\omega_i$, with $i=1,...16$, that have been used for the measurements. The $j^{th}$ measurement can be written as ${\bf x}^{[j]} = (r_1^{[j]}, r_2^{[j]}, ..., r_{16}^{[j]})$ with $r_i^{[j]} = r (\omega_i, T^{[j]}, [O_2]^{[j]})$ and $i=1,...16$. Each measurement $j$ corresponds to a specific tuple of the oxygen concentration and temperature $(T^{[j]}, [O_2]^{[j]})$.

Summarizing, the conventional approach relies on the measurement of the temperature, which is then used to correct the parameters of the analytical model used to calculate the oxygen concentration $\left[O_2\right]$ from the measured quantity, the phase shift $\theta$ of Equation \ref{theta_full}. The inadequate determination of the luminophore temperature is one of the major sources of error in an optical oxygen sensor.

The neural network proposed in this work defies the difficulties described above by simultaneously predicting both the oxygen concentration and the temperature using 16 values of $r (\omega, T, [O_2])$ evaluated at a discrete set of sixteen values of $\omega$.

\subsection{Data Generation}

To have a large enough dataset to train and test the neural networks, synthetic data were used. 
The model described by Equation (\ref{theta_full}) was chosen to create the data, being as simple as possible but still capable to describe experimental observations.
The values of the parameters for the synthetic data were determined from measurement performed under varying oxygen concentration and temperature conditions. For details on the samples and setup used for the determination of all the parameters the reader is referred to \cite{Michelucci2019}.

The synthetic data consist of a set $S$ of $m=25000$ observations using oxygen concentration values uniformly distributed between 0 $\%$ air and 100 $\%$ air and five temperatures $5, 15, 25, 35$ and $45  ^\circ C$. Please note that in the following, the concentration of oxygen is be given in $\%$ of the oxygen concentration of dry air and indicated with $\%$ air. This means that 100 $\%$ air corresponds to 20 $\%$ vol $O_2$. 
The $m$ data were split randomly in a training dataset containing 80 $\%$ of the data ($|S_{train}|=20000$), used to train the network, and a development dataset containing 20 $\%$ of the data ($|S_{test}|=5000$), used to test the generalisation efficiency of the network when applied to unseen data.

Typically when training neural network models, it is important to check if we are in a so-called overfitting regime. The essence of overfitting is to have unknowingly extracted some of the residual variation (i.e., the noise or errors) as if that variation represented an underlying model structure \cite{Burnham}. In the case discussed in this work, with increasing network complexity, the network will never go into such a  regime since the development dataset is a perfect representation of the training dataset. This leads to almost identical metric values for the $MAE$ for both $S_{train}$ and $S_{dev}$, regardless of the network architecture effective complexity. This is what we observed while checking the metrics on the two different dataset $S_{train}$ and $S_{dev}$. Overfitting becomes relevant when dealing with real measurements and not synthetic data.

\section{Results and Discussion}
\label{Results}

As described in Section \ref{Luminescence}, the applied problem investigated in this work is a complex one since the two quantities to be extracted from the data ($[O_2]$ and $T$) depend from the input in different ways. It is therefore not obvious that is possible to build a model which is able to predict both $[O_2]$ and $T$ at the same time with good accuracy.

\begin{figure}[hbt]
\centering
\includegraphics[width=15.5 cm]{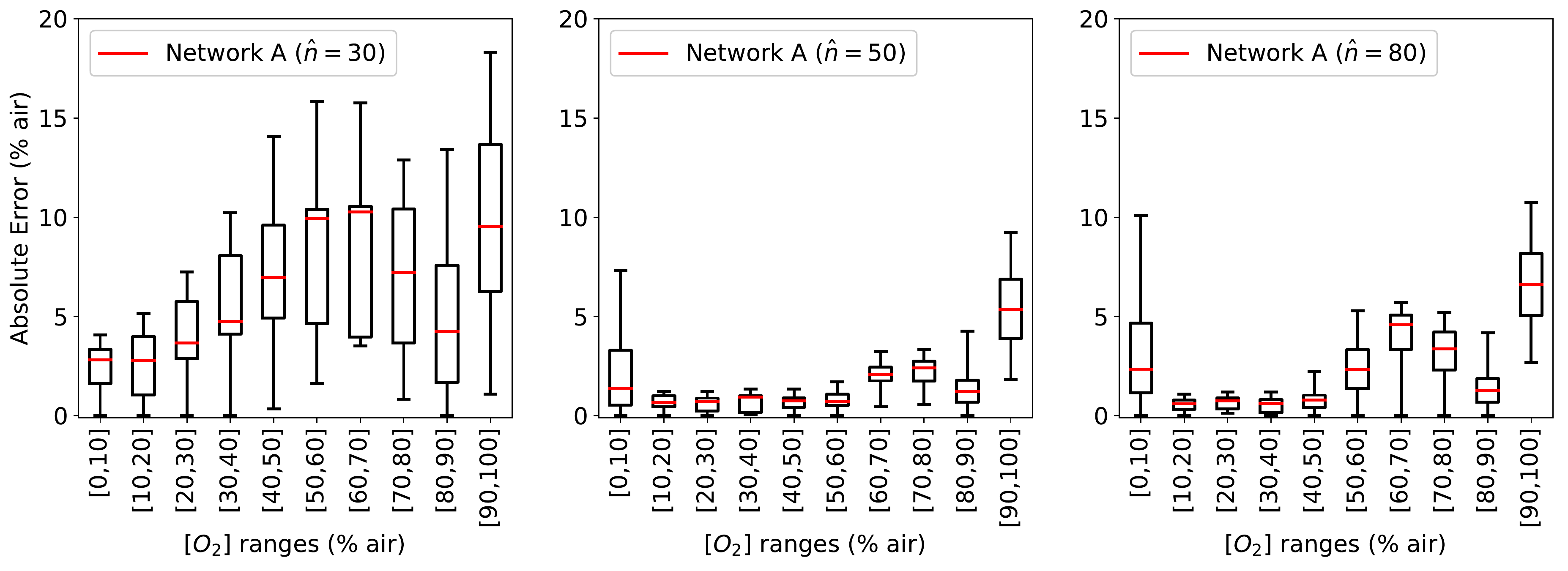}
\caption{Absolute error $AE_{[O_2]}$ in the prediction of the $O_2$ concentration for the different concentration ranges using network A. Left: 30 neurons per hidden layer; middle: 50 neurons per hidden layer, right: 80 neurons per hidden layer.}
\label{AE_A}
\end{figure}

The fist network investigated is the simple FFNN A described in Section \ref{NetworkA}. For this network, the number of neurons was progressively increased ($\hat n=10, 30, 50, 80$) to study how $AE_{[O_2]}$ and $AE_T$ are affected by an increasingly complex network and to determine if it is possible to obtain a good prediction. The calculated $AE_{[O_2]}$ for different $O_2$ concentrations were grouped in bins of 10 \% air for a clearer illustration and are shown in Figure \ref{AE_A} as a box plot, where the median is visible as a red line. In all the boxplots in this paper the central box is the interquartile range and contains the middle 50\% of the results, while the whiskers indicates the minimum and maximum of all the data \cite{boxplots}.

As it can be seen in Figure \ref{AE_A}, the results are quite poor if $\hat n=30$ (results for $\hat n=10$ are comparable to those with $\hat n=30$ and are not shown here). $AE_{[O_2]}$ can assume values as big as 18 \% air, with a broad distribution. Increasing the number of neurons in the hidden layers to $\hat n=50$ improves the prediction, reducing both the median and the distribution. A further increase to $\hat n=80$, however, does not result in better a prediction, showing the limits of this architecture to capture the details of the physical system.

\begin{figure}[hbt]
\centering
\includegraphics[width=15.5 cm]{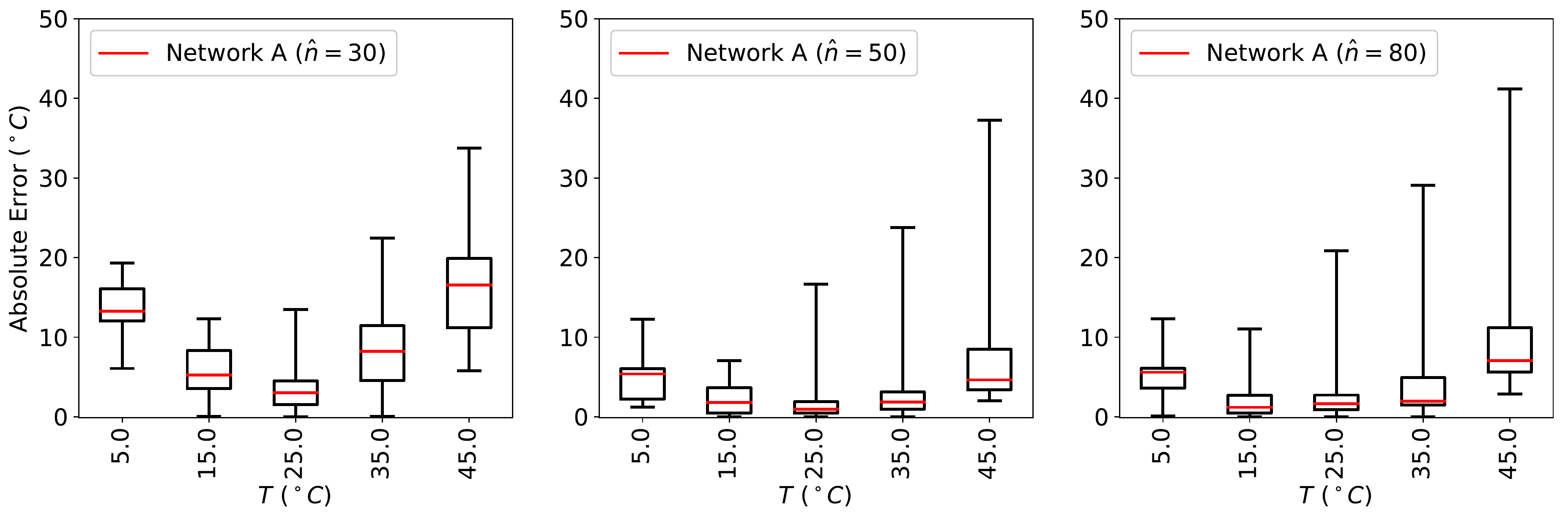}
\caption{Absolute error $AE_T$ in the prediction of $T$ for the different temperatures using network A. Left: 30 neurons per hidden layer; middle: 50 neurons per hidden layer, right: 80 neurons per hidden layer.}
\label{AET_A}
\end{figure}

The results for the prediction of the temperature for the same three networks are shown in Figure \ref{AET_A}. Also $AE_T$ improves initially by increasing the number of neurons to $\hat n=50$, but does not get any better when the number of neurons is further increased to $\hat n=80$. The boxplots of Figure \ref{AE_A} and Figure \ref{AET_A} show that $AE_{[O_2]}$ and $AE_T$ can assume quite high values, therefore demonstrating how the model is not really able to make a prediction with an accuracy that may be used in any commercial application.

The performance of the three FFNN of type A can be summarized calculating the $MAE$ as defined in Equation \ref{MAE}. The results are listed in Table \ref{TableMAE_A}. Consistently with what previously observed for the absolute error, the best network performance is obtained with $\hat n=50$, achieving a mean absolute error of $MAE_{[O_2]}=1.7 \ \%$ air and $MAE_T=3.3 \ ^\circ C$.  

\begin{table}[hbt]
\centering
\caption {\bf Summary of the performance for the FFNNs A}
\renewcommand*{\arraystretch}{1.2}
\begin{tabular}{ccc}
  \smallskip 
 $\hat n$ & $MAE_{[O_2]}$ & $MAE_{[T]}$  \\ 
\hline
30 & 6.0 \% air & 9.3 $^\circ C$\\
50 & 1.7 \% air & $ 3.3 \ ^\circ C$\\  
80 &  2.3 \% air & $4.3 \ ^\circ C$\\
 \end{tabular}
\label{TableMAE_A}
\end{table}

\begin{figure}[hbt]
\centering
\includegraphics[width=15.5 cm]{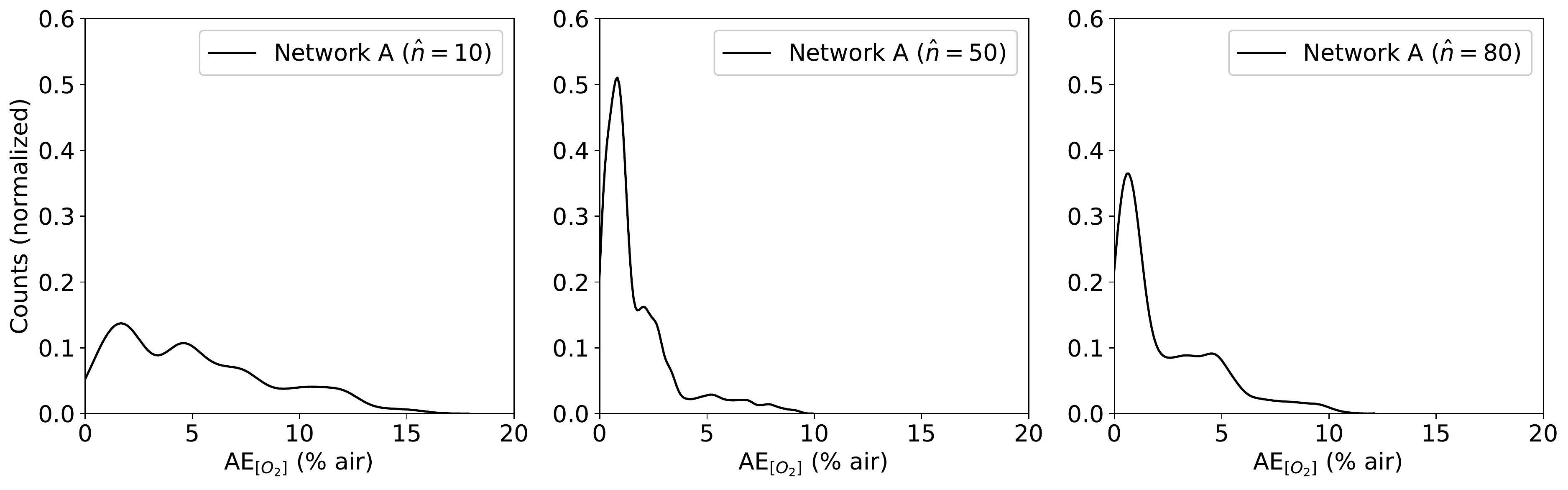}
\caption{Kernel density estimation for $AE_{[O_2]}$ with network A. Left: 30 neurons per hidden layer; middle: 50 neurons per hidden layer, right: 80 neurons per hidden layer.}
\label{dist1}
\end{figure}

\begin{figure}[h!]
\centering
\includegraphics[width=15.5 cm]{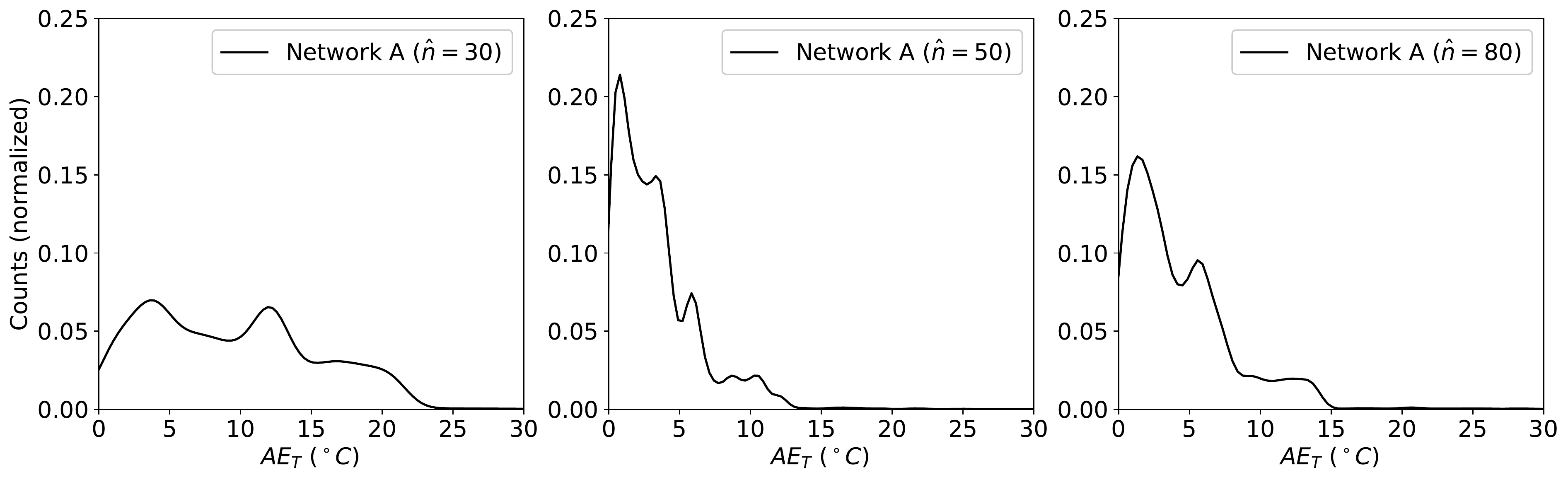}
\caption{Kernel density estimation for $AE_T$ for network A. Left: 30 neurons per hidden layer; middle: 50 neurons per hidden layer, right: 80 neurons per hidden layer.}
\label{dist2}
\end{figure}

For a practical application, the probability density distribution of the $AE$s for both parameters represent a much fundamental quantity since it carries information on the probability of the network to predict the expected value. 
For this reason, the  kernel density estimate (KDE) of the distributions of the $AE$s was used for analysis. KDE is a non-parametric algorithm to estimate the probability density function of a random variable by inferring the population distribution based on a finite data sample \cite{Hastie}. For the plots we have used a Gaussian Kernel and a Scott bandwidth estimation using the seaborn Python package \cite{seaborn}.
The results for $AE_{[O_2]}$ and $AE_T$ for the three variations of FFNN A are shown in Figure \ref{dist1} and \ref{dist2}, respectively.

From Figure \ref{dist1} and \ref{dist2} can clearly be seen that increasing the number of neurons helps at the beginning. A further increase in $\hat n$ does not produce an improvement in prediction quality, on the contrary it gets worse. These results indicate that this simple FFNN can extract at the same time the two quantities with an accuracy which is at best poor and at worst unusable.

\begin{figure}[hbt!]
\centering
\includegraphics[width=15.5 cm]{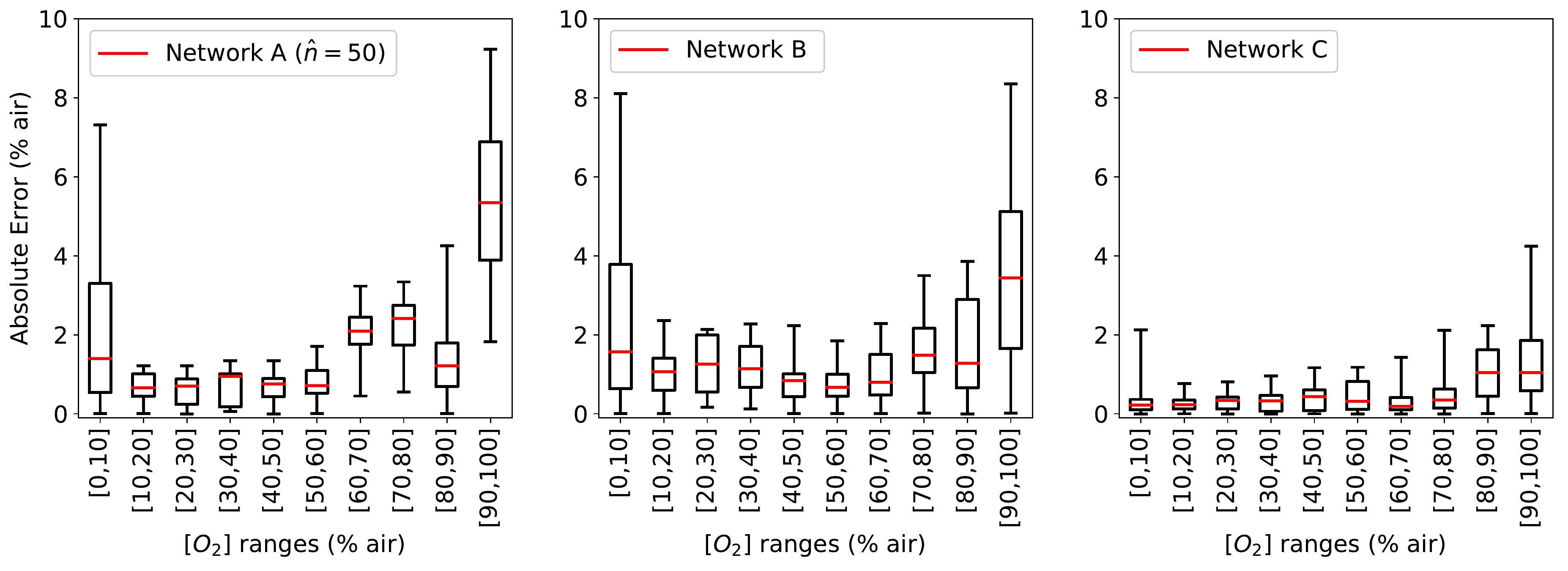}
\caption{Absolute error in the prediction of the $O_2$ concentration for the different concentration ranges using network A, B, and C. Left: Network A with 50 neurons per hidden layer; middle: network B, right: network C.}
\label{boxplot1}
\end{figure}

Networks B and C try to address this problem by adding, as described in previous sections, respectively one and two branches after the last hidden layer in network A. The results of the prediction from the networks B and C are then compared to those from network A with  $\hat n = 50$. Figure \ref{boxplot1} shows the calculated $AE_{[O_2]}$ for the three networks for the same $[O_2]$ intervals as before as a box plot, where the median is visible as a red line.

As it can be seen from Figure \ref{boxplot1}, the error in the prediction of network B is similar to that of network A. However, $AE_{[O_2]}$ is significantly improved when using network C. The additional branch in network C compared to network B clearly make the predictions much more accurate and, more importantly, much less spread around the median.

\begin{figure}[hbt]
\centering
\includegraphics[width=15.5 cm]{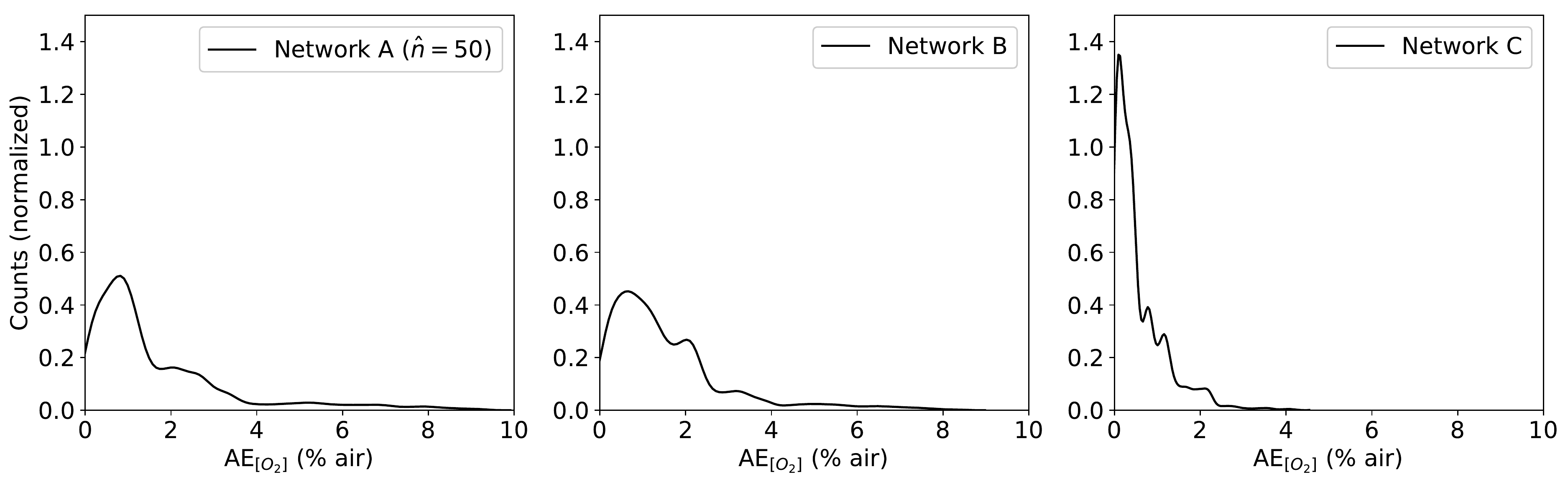}
\caption{Kernel density estimation for $AE_{[O_2]}$ for networks A (left), B (middle), and C (right).}
\label{dist3}
\end{figure}

The distribution of the $AE_{[O_2]}$ is better illustrated by plotting the KDE (Figure \ref{dist3}). The results indicate that the distribution assumes much smaller values and is peaked around zero for network C, in contrast with network A and B that have a quite wide tail that propagates toward higher values, reaching values as high as 10 \% air for network A and 8 \% air for network B.

\begin{figure}[hbt]
\centering
\includegraphics[width=15.5 cm]{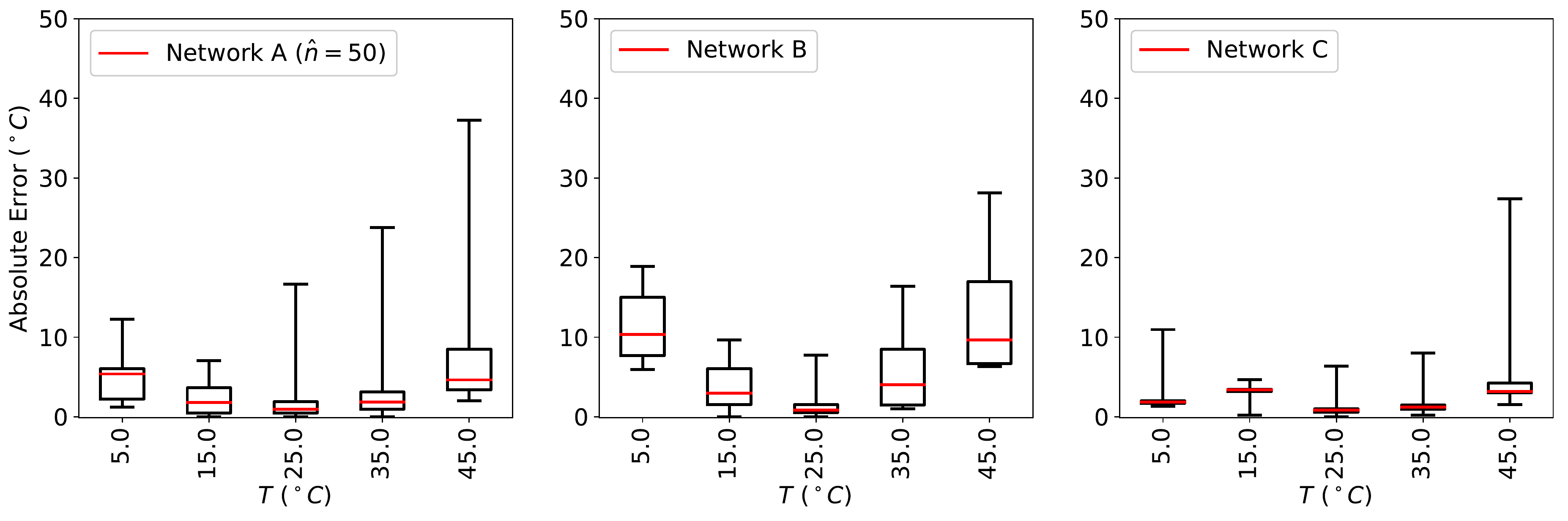}
\caption{Absolute error in the prediction of the temperature using network A, B, and C. Left: Network A with 50 neurons per hidden layer; middle: network B, right: network C.}
\label{boxplot2}
\end{figure}

Finally, the results of the same analysis for the prediction of the temperature are shown in Figure \ref{boxplot2}. Here  the calculated $AE_T$ for the same three networks is shown as a box plot, where the median is visible as a red line.
As it can be seen from Figure \ref{boxplot2}, $AE_T$ is much more concentrated around the median when using network C. 
These results indicate that the prediction of the temperature is substantially improved when using this network.

\begin{figure}[hbt]
\centering
\includegraphics[width=15.5 cm]{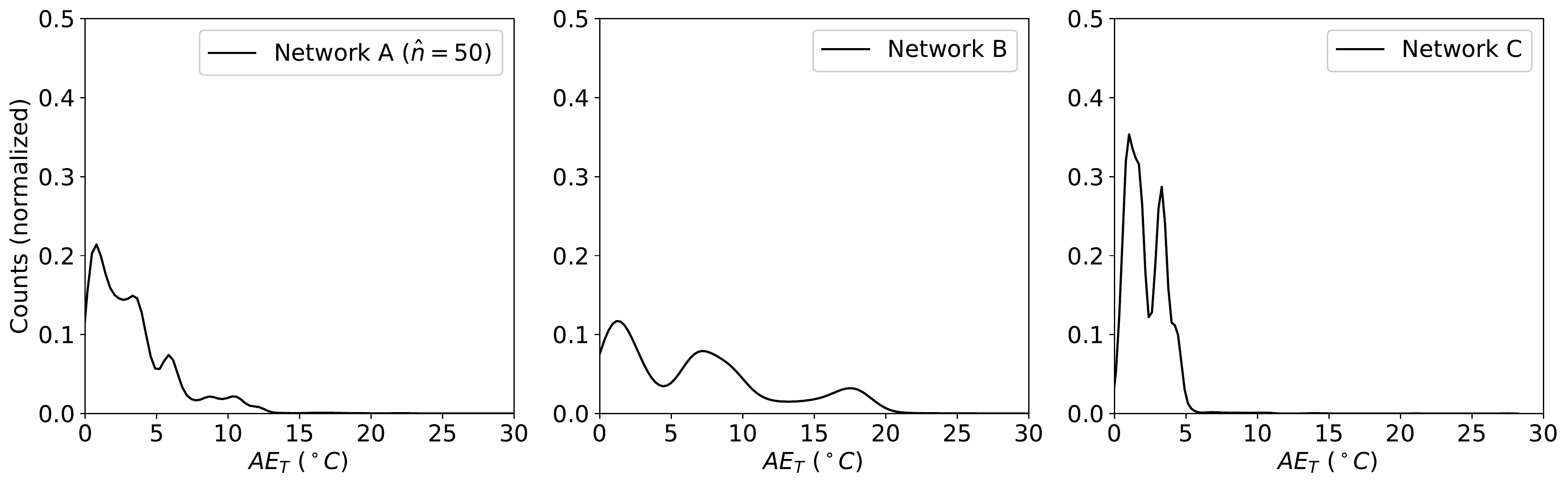}
\caption{Kernel density estimation for $AE_T$ for networks A (left), B (middle), and C (right).}
\label{dist4}
\end{figure}

The distribution of the $AE_T$ using the KDE is shown in Figure \ref{dist4}. Thanks to the additional task-specific hidden layer of the network C compared to network B, the KDE is higher and peaked around zero, with practically no contributions above $5^\circ C$.

Finally, the performance of the three neural networks are be summarized by calculating the $MAE$ as defined in Equation \ref{MAE} for the oxygen concentration and the temperature prediction. The results are listed in Table \ref{TableMAE_ALL}. The network C outperforms all the other networks analyzed in predicting both $[O_2]$ and $T$, achieving a mean absolute error of only 0.5 \% air for the oxygen concentration and of $2.2 \ ^\circ C$ for the temperature.  

\begin{table}[hbt]
\centering
\caption {\bf Summary of the performance for the three types of neural networks}

\begin{tabular}{ rcc}
\smallskip 
 Network & $MAE_{[O_2]}$ & $MAE_{[T]}$  \\ 
 \hline
 Network A ($\hat n = 30$) & 6.0 \% air & 9.3 $^\circ C$\\ 
 Network A ($\hat n = 50$) & 1.7 \% air & 3.3 $^\circ C$\\  
 Network A ($\hat n = 80$) & 2.3 \% air  & 4.3 $^\circ C$\\
 Network B & 1.5 \% air &   6.5 \ $^\circ C$  \\
 Network C & 0.5\% air  &  2.2 \ $^\circ C$ 
\end{tabular}
\label{TableMAE_ALL}
\end{table}

The results of Table \ref{TableMAE_ALL} show that a simple FFNN as network A is not suitable to extract the two quantities of interest at the same time with good accuracy, since it is not flexible enough. The reason is that the two predicted quantities will depend on the same set of features generated by the hidden layers of network A. When network A tries to learn better weights to predict, for example, the temperature, these will, however, influence also the $[O_2]$ prediction and vice-versa. So the common set of weights that are learned can not be optimized for each quantity separately at the same time.
The MTL network B tries to address this problem with a separate branch of task-specific layers for $[O_2]$. The tests show however that this architecture is only marginally better for the prediction of $[O_2]$ and even worse for the prediction of $T$. This is probably due an insufficient flexibility of the network and shows that even if only one parameter were of interest, e.g., $[O_2]$ one single additional branch is not sufficient. 
A significant improvement is achieved with the MTL network C: the two task-specific branches give the network the flexibility of learning a set of weights (the ones in the branches) specific to each quantity, therefore achieving exceptionally good predictions on both $[O_2]$ and $T$. 
Note that in this work the hyper-parameter tuning \cite{Michelucci2017} for each network was not performed since the goal is not to achieve the lowest possible $MAE$s but rather to demonstrate the advantages and potential of MTL compared to classical FFNN approaches. For the implementation in a measuring instrument, therefore, a further phase of parameter tuning specifically dependent on the application would be needed.
 
An interesting question is what is the mutual influence of the branches in network C when the loss weights $\alpha_i$ are varied. To answer this question, a study was performed with various values of the global cost function weights. The results are shown in Table \ref{tableweights}.

\begin{table}[hbt]
\centering
\caption {\bf Summary of the performance for network C with various loss weights}
\begin{tabular}{ rcccc}
\smallskip 
 $\alpha_1$ & $\alpha_2$ & $\alpha_3$  & $MAE_{[O_2]}$ & $MAE_{[T]}$\\ 
 \hline
0.3 & 5.0 & 5.0 & 0.54 \% air & 2.2 \ $^\circ C$\\ 
0.3 & 5.0 & 15.0 & 0.61 \% air & 2.35 \ $^\circ C$\\ 
0.3 & 5.0 & 25.0 & 0.89 \% air & 2.32 \ $^\circ C$\\ 
0.3 & 1.0 & 5.0 & 0.58 \% air & 2.25 \ $^\circ C$\\ 
0.3 & 15.0 & 5.0 & 0.94 \% air & 2.67 \ $^\circ C$\\ 
0.3 & 25.0 & 5.0 & 0.96 \% air & 2.55 \ $^\circ C$\\ 
\end{tabular}
\label{tableweights}
\end{table}

By increasing progressively the weight for the temperature branch, $\alpha_3$, the MAE$_{[T]}$ is not reduced further and appears rather insensitive to $\alpha_3$. However, $MAE_{[O_2]}$ increases slightly, since the higher values of $\alpha_3$ shift the relative importance of the tasks the network is trying to learn.
Increasing the weight for the oxygen branch $\alpha_2$ negatively affects the oxygen prediction since $MAE_{[O_2]}$ increases slightly. The reason why this is happening is that the $\alpha_2$ is becoming much bigger than $\alpha_1$. This shows that for the prediction of the oxygen concentration both the branches predicting $T$ and $[O_2]$ at the same time and the one predicting $[O_2]$ are important: neglecting one will make the other works less efficiently. 
The temperature, on the other had, is predicted almost with the same kind of accuracy independently of the weights $\alpha_2$, indicating that the temperature branch is not dependent from the $[O_2]$ branch.

\section{Conclusions}

In this work, different neural networks architectures were investigated to solve the problem of extracting multiple separate physical quantities at the same time from a single dataset. This type of multi-dimensional regression problems in physics can be challenging or impossible to solve if the mathematical models describing the functional dependence of  the dependent variable from a set of independent variables are too complex or unknown.
The proposed approach consists in using neural network MTL architectures, which are characterized by a common set of layers and then task-specific layers for each quantity to be determined. Thanks to the additional task-specific hidden layers, this type of networks can be trained to perform better than conventional FFNNs when the quantities to be predicted are characterized by a significant difference in physical behavior.

The approach is demonstrated by applying it to oxygen luminescence sensing application. The conventional methods rely on a separate temperature determination which is then used as input to correct the extraction of the oxygen concentration from a dataset. This work demonstrates how it is possible to extract from a single dataset of phase shift measurements both the oxygen concentration and the temperature of the medium. The distributions of $AE_{[O_2]}$ and $AE_T$ are significantly narrower and much more concentrated around zero with the proposed MTL network (type C), as compared to FFNNs without specific and dedicated layers for each $[O_2]$ and $T$. With the latter network the predictions are only based on common features (the ones generated by the common layers) that fail to be flexible enough to describe both $[O_2]$ and $T$. The results indicate that from one single measurement, it is possible to determine two physically different quantities, one of which is dependent from the other.
To the best of the authors' knowledge this is the first time that more than one parameter (here $[O_2]$ and $T$) are extracted using a single luminophore by a single measurement channel under constant conditions. The implication is that a sensor using the proposed approach could be able to extract much more information from the measurements than one based on conventional analytical modeling.

This work aims to open the road to new ways of extracting multiple  physical quantities from a common set of data at the same time to achieve consistent results that are both  accurate and stable. The described approach is relevant for many practical applications in sensor science and demonstrates that MTL architectures have the potential of revolutionizing the approach to non-linear multi-dimensional regression.

\authorcontributions{conceptualization, Umberto Michelucci and Francesca Venturini; methodology, Umberto Michelucci and Francesca Venturini; software, Umberto Michelucci; writing, Umberto Michelucci and Francesca Venturini; physics model and examples, Francesca Venturini}



\conflictsofinterest{The authors declare no conflict of interest.} 

\abbreviations{The following abbreviations are used in this manuscript:\\

\noindent 
\begin{tabular}{@{}ll}
FFNN & Feed-forward neural networks \\
MTL & Multi-task learning\\
MSE &  Mean square error \\
AE & Absolute error \\
MAE &  Mean average error \\
KDE & Kernel density estimate \\
\end{tabular}}

%


\reftitle{References}





\end{document}